\pdfoutput=1

\documentclass[11pt]{article}

\usepackage[final]{acl}

\usepackage{times}
\usepackage{latexsym}

\usepackage[T1]{fontenc}

\usepackage[utf8]{inputenc}

\usepackage{microtype}

\usepackage{inconsolata}

\usepackage{graphicx}
\usepackage{mathrsfs}
\usepackage{booktabs}    
\usepackage{amssymb}
\usepackage{amsthm}
\usepackage{subcaption}
\usepackage{mathtools}
\usepackage{multirow}
\usepackage{wrapfig}
\usepackage{xspace}

\usepackage{algorithm}
\usepackage{algorithmic}
\usepackage[absolute,overlay]{textpos}
\usepackage[most]{tcolorbox}

\newcommand{\mymethod}{\textsc{SaMuLe}\xspace}

\title{\mymethod: Self-Learning Agents Enhanced by Multi-level Reflection}

\author{Yubin Ge\textsuperscript{$1$}, Salvatore Romeo\textsuperscript{$1$}, Jason Cai\textsuperscript{$1$}, Monica Sunkara\textsuperscript{$1$}, Yi Zhang\textsuperscript{$1$} \\
\textsuperscript{$1$}AWS AI Labs \\
\texttt{\{yubinge, romeosr, cjinglun, sunkaral, yizhngn\}@amazon.com}}

\begin{document}
\maketitle
\begin{abstract}
    Despite the rapid advancements in LLM agents, they still face the challenge of generating meaningful reflections due to inadequate error analysis and a reliance on rare successful trajectories, especially in complex tasks. In this work, we propose \mymethod, a new framework for self-learning agents powered by a retrospective language model that is trained based on Multi-Level Reflection Synthesis. It first synthesizes high-quality reflections across three complementary levels: Single-Trajectory Learning (micro-level) for detailed error correction; Intra-Task Learning (meso-level) to build error taxonomies across multiple trials of the same task, and Inter-Task Learning (macro-level) to extract transferable insights based on same typed errors from diverse task failures. Then we fine-tune a language model serving as the retrospective model to generate reflections during inference. We further extend our framework to interactive settings through a foresight-based reflection mechanism, enabling agents to proactively reflect and adapt during user interactions by comparing predicted and actual responses. Extensive experiments on three challenging benchmarks—TravelPlanner, NATURAL PLAN, and Tau-bench—demonstrate that our approach significantly outperforms reflection-based baselines. Our results highlight the critical role of well-designed reflection synthesis and failure-centric learning in building self-improving LLM agents.
\end{abstract}

\section{Introduction}

Modern AI agents~\cite{yao2023react,fourney2024magentic,expel,self_ev} increasingly rely on Large Language Models (LLMs) as their core reasoning engines, empowering them to understand natural language instructions, reason through multi-step processes, and interact with external environments \cite{ge2025tremu, shen2025optimizing}. 
These models typically undergo pre-training on extensive text corpora, endowing them with unprecedented accuracy in predicting the next token given some context \cite{ge2023supervised}.
However, LLM agents remain fundamentally limited in their ability to autonomously improve from experience, particularly in complex, failure-prone environments~\cite{ji2024testing,wang2024learning}. This bottleneck severely restricts their utility in scenarios where learning from failures is critical for long-term success.

Existing experiential learning methods for LLM agents suffer from several core limitations. First, they sometimes fail to generate meaningful and actionable reflections after failures due to inadequate error analysis mechanisms. For example, Reflexion~\cite{shinn2023reflexion} shows limited improvements on complex benchmarks like TravelPlanner~\cite{xie2024travelplanner}, as it lacks the capacity to deeply diagnose failure causes and therefore produces general and useless strategies for correction. 

Second, many sophisticated prompting-based methods, such as Expel~\cite{expel}, depend heavily on successful trajectories as learning signals and demonstrate the benefits of utilizing knowledge \cite{ge2021baco}. This reliance on successful trails makes them impractical in real-world settings where task success is rare and failures are far more prevalent. 
By failing to harness the rich information embedded in unsuccessful attempts, these methods miss critical opportunities for learning and adaptation, leading to poor generalization in complex environments.

Besides, advanced approaches like Retroformer~\cite{yao2024retroformer} and CTRL~\cite{xie2025teaching}, which use reinforcement learning (RL) and rewards based on task improvements from synthesized reflections, are sensitive to the quality of the synthesized reflections. In complex tasks where existing reflection algorithms struggle to produce informative and accurate feedback, these RL-based methods suffer from learning meaningless policies.

In this work, we propose \mymethod, a framework for \textbf{S}elf-learning \textbf{A}gents enhanced by \textbf{MU}lti-\textbf{LE}vel reflections that unlocks the learning potential of past trajectories through training a retrospective model. Specifically, we first design a Multi-Level Reflection Synthesis to synthesize high-quality reflection data across three complementary levels of granularity: \textbf{Single-Trajectory Learning (Micro-Level)}: Analyzing individual failed trajectories against reference plans to identify immediate errors and generate targeted corrective strategies; \textbf{Intra-Task Learning (Meso-Level)}: Examining multiple trajectories from the same task query to categorize error types and build an error taxonomy, enabling richer, pattern-based feedback; \textbf{Inter-Task Learning (Macro-Level)}: Clustering similar errors across diverse task queries to derive high-level, transferable insights that improve future decision-making across tasks. After merge the reflections from different levels as the final target reflection, we then train a language model through SFT, which dynamically generates trajectory-specific reflections for the agent during inference.

We further extend to the interactive setting, where agents decide whether and when to reflect during interactions with users. We introduce foresight-based reflection that compares the agent’s predicted user response with the actual response at each turn. When the true response is beyond expectation, the agent triggers a reflection step and adds the generated feedback into its ongoing interaction, enabling real-time correction and adaptation.

We extensively evaluate our approach across three challenging benchmarks: TravelPlanner~\cite{xie2024travelplanner}, NATURAL PLAN~\cite{zheng2024natural}, and Tau-bench~\cite{yaotau}. Experimental results consistently demonstrate that our method substantially outperforms existing reflection-based baselines, particularly in complex, failure-dense environments. Notably, we show that even with simple supervised fine-tuning, our retrospective model—trained on multi-level synthesized reflections—achieves superior performance compared to advanced methods relying on RL. These findings underscore the critical role of well-designed reflection synthesis in efficient learning from past trajectories without the need for costly or unstable RL training. Moreover, the strong results achieved in interactive settings highlight the generalizability of our framework and its ability to support real-time adaptive learning. By leveraging failures as rich learning opportunities and introducing structured reflection at multiple levels, our work represents a significant step toward building more resilient, adaptive, and self-improving agents.

Our contributions are summarized as follows:
\begin{itemize}
    \item We propose \mymethod, a new self-learning framework that introduces to training a retrospective model based on Multi-Level Reflection Synthesis spanning from micro-level to macro-level analysis.
    \item We introduce foresight-based reflection to extend our framework to interactive scenarios, enabling agents to proactively reflect and adapt during user interactions by comparing predicted and actual user responses.
    \item Extensive experiments on TravelPlanner, NATURAL PLAN, and Tau-bench show that our approach significantly outperforms existing reflection-based baselines, particularly in complex benchmarks, highlighting the value of failure-centric learning and multi-level reflection for building self-improving agents.
\end{itemize}

\section{Related Work}



\textbf{Prompt-based self-reflection and self-learning.} One foundational self-reflective capabilities for Gen AI agents was established by \citet{shinn2023reflexion}, who introduced Reflexion, a framework enabling language agents to learn from verbal reinforcement through self-reflection. Building on this foundation, research has made significant advances in developing sophisticated self-reflective capabilities through various prompting strategies~\cite{selfreflection_makes_2024,selfreflection_in_2024_2405,iteration_of_2024_2409,selfcontrast_better_2024_2401}. \citet{selfreflection_makes_2024} demonstrated that self-reflection can enhance model safety and reduce bias while maintaining ideological neutrality. Further validation came from \citet{selfreflection_in_2024_2405}, who documented substantial improvements in problem-solving performance through structured reflective processes. To address the challenge of reflection stability, \citet{selfcontrast_better_2024_2401} developed Self-Contrast, introducing a novel approach that explores and contrasts diverse solving perspectives. The field has rapidly expanded into experiential learning, with \citet{zhao2024expel} introducing EXPEL for experiential learning in LLM agents with the support of autonomous experience extraction from training tasks, while \citet{gao-etal-2024-self-evolving} advanced the state-of-the-art with a self-evolving framework enabling life-long experiential learning. Recent work has also explored novel directions, including \citet{qian-etal-2024-experiential}'s co-learning framework for software development, \citet{iteration_of_2024_2409}'s Inner Dialogue framework for autonomous reasoning, and \citet{gödel_agent_2024_2410}'s self-referential framework for recursive self-improvement. 

\textbf{Post-training powered by self-reflection.} The integration of self-reflection into post-training optimization has emerged as a powerful approach for enhancing LLM capabilities~\cite{yao2024retroformer,improving_retrospective_2025_2503,webrl_training_2024_2411,selfrewarding_correction_2025_2502,online_preferencebased_2024_2412,experiential_explanations_2022_2210}. \citet{yao2024retroformer} pioneered this direction with Retroformer, introducing policy gradient optimization for retrospective learning in language agents. \citet{improving_retrospective_2025_2503} designed a sophisticated two-stage optimization process combining imitation learning with reinforcement learning, which enhances the data efficiency and training stability. For web-based applications, \citet{webrl_training_2024_2411} developed WebRL, demonstrating how self-evolving online curriculum learning can substantially improve open-source LLM performance. Another promising direction has emerged in self-rewarding mechanisms, exemplified by \citet{selfrewarding_correction_2025_2502}'s work on mathematical reasoning, where models simultaneously generate and evaluate their reasoning without external feedback. The frontier of autonomous self-improvement has been further pushed by \citet{scaling_autonomous_2025_2502}, who established a comprehensive framework for scaling autonomous agents through automatic reward modeling. Additional advances include \citet{online_preferencebased_2024_2412}'s online preference-based reinforcement learning approach and \citet{experiential_explanations_2022_2210}'s work on experiential explanations for reinforcement learning, both contributing to more robust self-reflection mechanisms.

\section{Methodology}
\subsection{Problem Setup}

\begin{figure*}[ht]
\centering
\includegraphics[width=0.85\linewidth]{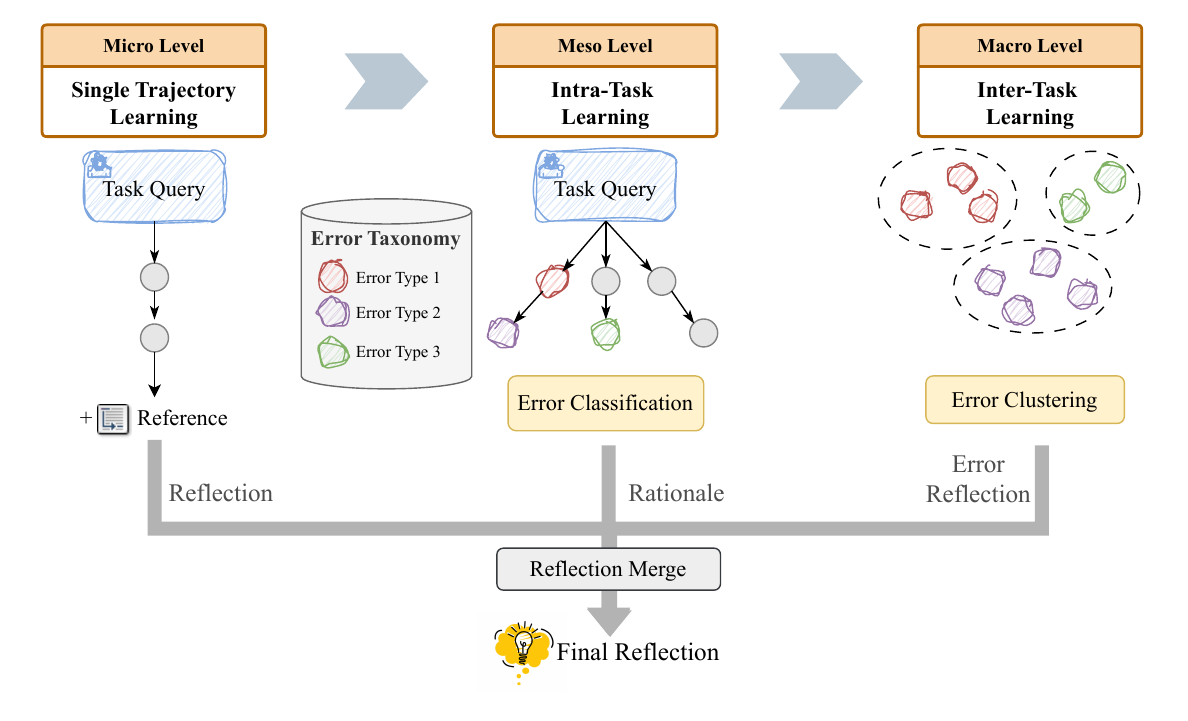}
    \caption{The overview of our proposed Multi-level Reflection Synthesis. The circles represents the steps in a trajectories and the sketched circles in different color, like in red, purple and green, indicating the steps are classified as different types of errors.} 
    \label{fig:multi_level}
\vspace{-0.5cm}
\end{figure*}

\subsubsection{General Setting}
We consider a setting where an agent performs complex tasks over a sequence of time steps. Following a standard machine learning setup, we assume access to training data $\mathcal{D}_{\text{train}} = \{(x^{(i)}_{\text{train}}, y^{(i)}_{\text{train}})\}_{i=1}^N$, which allows the agent to learn via task exploration. Each input $x_{\text{train}}^{(i)}$ includes an instruction, a task query, and relevant background information, while the corresponding output $y_{\text{train}}^{(i)}$ represents a valid response, such as a human-annotated plan to the input query. We consider $y_{\text{train}}^{(i)}$ as a reference. Particularly, in agentic task completion, $y_{\text{train}}^{(i)}$ does not need to be the unique ground truth but is considered a valid solution to $x_{\text{train}}^{(i)}$. During inference, the agent receives a test input $x_{\text{test}}^{(i)}$ and generates a predicted output $\hat{y}_{\text{test}}^{(i)}$ for evaluation.

\subsubsection{Reflection Setting}
We explore two settings for reflection:

\textbf{Non-Interactive}: We follow the self-reflection setup from Reflexion \cite{shinn2023reflexion}, where the agent reflects only after completing a trial. If the trial fails, the corresponding trajectory is reviewed to generate a reflection. This reflection is then appended to the input for the next trial.
    
\textbf{Interactive}: In real-world scenarios, agents often interact with users through multi-turn conversations to complete tasks. In this setting, the agent may not be able to try the task multiple times and therefore cannot defer reflection until after the task concludes. Instead, it must dynamically decide whether to reflect during the interaction.

\subsection{\mymethod}
We propose a framework for self-learning LLM agents by training a retrospective language model that is subsequently used to refine the LLM agent through reflective feedback on its past trajectories.

\subsubsection{Stage I: Multi-Level Reflection Synthesis}
Prior studies show that learners who engage in self-explanation while studying examples perform better and develop more robust knowledge structures than those who do not. A key reason is that as learners explain individual steps (concrete level), they often invoke domain principles or rules (abstract level) to justify why a step is correct or incorrect. This bridges the gap between specific feedback and general understanding \cite{wang2022learning}. In fact, having to explain why an error is an error forces learners to integrate the relevant conceptual knowledge with the procedure \cite{tulis2016learning}. 

Motivated by these findings, we introduce a multi-level reflection framework that synthesizes reflections from the micro-level (detailed, instance-specific) to the macro-level (general, conceptual). Our design aligns with Kolb’s experiential learning model, which describes a learning cycle involving concrete experiences, reflective observation, and abstract conceptualization for effective knowledge acquisition \cite{mcleod2017kolb}. The overview is shown in Figure~\ref{fig:multi_level} and Algorithm \ref{alg:multi_level} (the detailed prompts are shown in Appendix $\S$ \ref{sec:multi_prompt}), and the process consists of the following steps:

\begin{algorithm}[ht]
\small
    \caption{Multi-Level Reflection Synthesis}
    \begin{algorithmic}
    \STATE \textbf{Initialize:} 
    \STATE Training data $\mathcal{D}_\text{train} \{(x^{(i)}_{\text{train}}, y^{(i)}_{\text{train}})\}_{i=1}^N$
    \STATE Trajectory pool $\mathcal{P} \leftarrow \{\}$
    \STATE Number of training tasks $N$
    \STATE Maximum retry number $\mathcal{K}$
    \STATE Error Taxonomy $\mathcal{E} \leftarrow \emptyset$
    \STATE \texttt{// Single Trajectory Learning}
    \FOR{task $i = 1$ to $N$}
        \STATE reflection \( r_{i,0} \leftarrow \text{""} \)
        \FOR{trial $k = 1$ to $\mathcal{K}$}
            \STATE trajectory \(\tau_{i,k} \leftarrow \text{LLM}_\text{ReAct}(x_\text{train}^{(i)}, r_{i, k-1}) \)
            \STATE \( \mathcal{P}[i] \leftarrow \mathcal{P}[i] \cup \tau_{i,k} \)
            \IF{ \( \texttt{fail}(\tau_{i,k}) \) }
                \STATE \( r_{i,k} \leftarrow \text{LLM}_\text{reflect}(\tau_{i,k}, y_{\text{train}}^{(i)})) \)
            \ENDIF
        \ENDFOR
    \ENDFOR
    
    \STATE \texttt{// Intra-Task Learning}
    \FOR{task $i=1$ to $N$}
        \STATE \(\mathcal{E} \leftarrow \mathcal{E} \cup \text{LLM}_{\text{error}}(\mathcal{E}, \tau_{i,1}, \dots, \tau_{i,\mathcal{K}}) \)
        \FOR{trial $k=1$ to $\mathcal{K}$}
            \STATE \(\epsilon_{i,k}, z_{i,k} \leftarrow \text{LLM}_{\text{error}}(\mathcal{E}, \tau_{i,k}))\)
        \ENDFOR
    \ENDFOR

    \STATE \texttt{// Inter-Task Learning}
    \FORALL{error $e \in \mathcal{E}$}
        \STATE grouped trajectory \(g_{\text{traj}} \leftarrow \emptyset\)
        \FORALL{trajectory $\tau_{i,k} \in \mathcal{P}$}
            \IF{error path $\epsilon_{i,k} \supset e$}
                \STATE \(g_{\text{traj}} \leftarrow g_{\text{traj}} \cup \tau_{i,k}\)
            \ENDIF
        \ENDFOR
        \STATE error reflection \(r_{e} \leftarrow \text{LLM}_\text{reflect}(g_{\text{traj}})\)
    \ENDFOR

    \STATE \texttt{// Reflection Merge}
    \FORALL{$\tau_{i,k} \in \mathcal{P}$}
        \STATE grouped error reflection \(g_e \leftarrow \emptyset\)
        \FORALL{error $e \in \mathcal{E}$}
            \IF{error path $\epsilon_{i,k} \supset e$}
                \STATE \(g_e \leftarrow g_e \cup r_e\)
            \ENDIF
        \ENDFOR
        \STATE final reflection \(r^\text{final}_{i,k} \leftarrow \text{LLM}_{\text{summ}}(r_{i,k}, z_{i,k}, g_e)\)
    \ENDFOR
    \end{algorithmic}
    \label{alg:multi_level}
\end{algorithm}
\vspace{-0.0cm}

\textbf{Single Trajectory Learning (micro-level)}: At the micro level, we follow Reflexion \cite{shinn2023reflexion} to iteratively retry each training task query at most $\mathcal{K}$ times. Specifically, at the $k$-th trial, given the current training instance $(x_\text{train}^{(i)}, y_\text{train}^{(i)})$ and the current reflection $r$ (initialized as an empty string), the agent attempts the task $x_\text{train}^{(i)}$ using ReAct \cite{yao2023react} as the base planning algorithm, resulting in a trajectory $\tau_{i,k} = \text{LLM}_\text{ReAct}(x_\text{train}^{(i)}, r)$. If the agent fails, we prompt the LLM to perform reflection. Inspired by prior work on contrastive learning from past successes and failures \cite{sun2023contrastive}, we instruct LLM to compare the failed trajectory with the reference output $y_{\text{train}}^{(i)}$, diagnoses the cause of failure, and generates a new, concise plan that addresses the identified issues as the reflection $r_{i,k} = \text{LLM}_{\text{reflect}}(\tau_{i,k}, y_{\text{train}}^{(i)})$. Then in the next retry, the agent augments its context with the updated reflection $r_{i,k}$ to attempt the task again.
    
\textbf{Intra-Task Learning (meso-level)}: This stage focuses on learning from multiple trajectories—both successful and failed—associated with the same task query. For each task $x_\text{train}^{(i)}$ we concatenate all attempted trajectories and ask the LLM to identify common failure patterns, incrementally constructing an error taxonomy $\mathcal{E}$. Specifically, the LLM first checks the existing taxonomy, appends any new error types it encounters, $\mathcal{E} = \mathcal{E} \cup \text{LLM}_{\text{error}}(\mathcal{E}, \tau_{i,1}, \dots, \tau_{i,\mathcal{K}})$. Then the LLM reviews each trajectory $\tau_{i,k}$ to label each action in the trajectories with error types (if applicable) to obtain an error path $\epsilon_{i,k}$, along with a rationale $z_{i,k}$ for the classification, i.e., $\epsilon_{i,k}, z_{i,k} = \text{LLM}_{\text{error}}(\mathcal{E}, \tau_{i,k})$.
    
\textbf{Inter-Task Learning (macro-level)}: At the macro level, we process each error type $e \in \mathcal{E}$ and group trajectories from different task queries that exhibit the error $e$ into a cluster $g_{\text{traj}}$. For each cluster, the LLM is prompted to generate a reflection that generalizes across tasks, identifying recurring failure modes and proposing strategies to mitigate these errors beyond individual trajectories, i.e., $r_e = \text{LLM}_{\text{reflect}}(g_{\text{traj}})$. Then each trajectory is paired with the error type reflection for the corresponding error types it contains.

\textbf{Reflection Merge}: We concatenate reflections at all levels for each trajectory $\tau_{i,k}$ and apply a final summarization step to produce a merged reflection $r^{\text{final}}_{i,k}$. This final reflection integrates both instance-specific feedback and generalized error patterns derived from related trajectories.

\subsubsection{Stage II: Retrospective Model Training}
Since our multi-level reflection framework depends on the reference output $y_{\text{train}}^{(i)}$, it cannot be directly applied during inference. To overcome this limitation, we train a smaller language model—referred to as the retrospective model—using the synthesized multi-level reflections.

We construct training examples by concatenating the instruction, task query, background information, and trajectory as the input, and using the corresponding synthesized reflection as the target output. The retrospective model is trained via supervised fine-tuning (SFT). During inference, it takes in an agent's trajectory and generates a reflection, thereby enabling reflective feedback without access to reference outputs.

\begin{algorithm}
\small
    \caption{Foresight-based Reflection}
    \begin{algorithmic}
    \STATE \textbf{Initialize:} 
    \STATE Actor LLM\textsubscript{ReAct}
    \STATE Self-reflection LLM\textsubscript{reflect}
    \STATE Current task $x^{(i)}$
    \STATE Maximum step number $\mathcal{H}$
    \STATE Initialize trajectory \(\tau_i \leftarrow \text{env.reset}()\)
    \STATE Initialize reflection \( r \leftarrow \text{""} \)
    \FOR{step $t = 1$ to $\mathcal{H}$}
        \STATE Action \(a_t \leftarrow \text{LLM}_\text{ReAct}(a_t| \tau_i, r )\)
        \STATE Predicted response \(\mathcal{R}_p \leftarrow \text{LLM}(\tau_i, a_t)\)
        \STATE True response \(\mathcal{R}_t, \texttt{done} \leftarrow \text{env.step}(a_t)\)
        \IF{$\text{LLM}_{\text{diff}}(\mathcal{R}_p, \mathcal{R}_t)$}
            \STATE \( r \leftarrow \text{LLM}_\text{reflect}(\tau_i)) \)
        \ENDIF
        \STATE \(\tau_i \leftarrow (a_t, \mathcal{R}_t)\)
        \IF{\texttt{done}}
            \STATE break
        \ENDIF
    \ENDFOR
    \end{algorithmic}
    \label{alg:foresight}
\end{algorithm}
\vspace{-0.2cm}

\subsubsection{Foresight-based Reflection}

Motivated by prior work that both foresight and reflection are essential for LLM-based Theory of Mind \cite{zhou2023far}, we introduce foresight-based reflection (shown in Algorithm \ref{alg:foresight}) to extend our framework to the interactive setting, where the agent engages in an interaction trial with the user.

In this setting, the agent compares its \textit{predicted} user response with the \textit{actual} response at each turn. Specifically, during each step, the agent first predicts the user's response $\mathcal{R}_p$ by prompting itself based on the current trajectory. Once the true user response $\mathcal{R}_t$ is observed, the agent is asked to compare $\mathcal{R}_p$ and $\mathcal{R}_t$, and then decides whether to reflect due to any significant deviation between expectation and reality. In such cases, the agent is prompted to perform reflection based on the current trajectory. This mechanism enables the agent to detect unexpected user behavior during interaction and adaptively revise its plan, improving alignment and task success in real-time settings.

\section{Experiments}
We introduce experimental details in this section and add implementation details in Appendix $\S$ \ref{sec:implementation}.

\subsection{Datasets}

\textbf{TravelPlanner} \cite{xie2024travelplanner} is a challenging benchmark focused on travel planning scenarios. It offers a comprehensive evaluation environment with nearly four million data records and 1,225 carefully curated planning queries, with each training instance paired with a reference plan. In our experiments, we adopt the \emph{sole-setting}, where relevant background information is provided directly, rather than the \emph{two-staged setting} that requires agents to actively search for information through tool usage. This choice is motivated by computational considerations: trajectories generated under the sole-setting can exceed 10k tokens, while those in the two-staged setting are even longer, rendering training computationally infeasible given our resources. The experiments are conducted on the validation set, as the complete test data for evaluation is not completely released and we are not able to perform reflection based on the task outcomes.

\textbf{NATURAL PLAN} \cite{zheng2024natural} assesses the planning capabilities of LLMs using complete information retrieved from real-world tools, including Google Flights, Google Maps, and Google Calendar. It emphasizes structured reasoning over tool outputs to generate feasible and coherent plans. Following prior work \cite{lee2025evolving}, we exclude Calendar Scheduling tasks, as these can be trivially solved through enumeration. We split the dataset into training and testing sets using an 80:20 ratio.

\textbf{Tau-bench} \cite{yaotau} is used to evaluate agent performance under interactive and non-interactive settings. It simulates conversations between a simulated user and an agent equipped with domain-specific APIs and policy guidelines.

\subsection{Baselines}

We compare our approach with the following reflection-based baselines:

\textbf{Fewshot}: For the setting that there is no environment that the agent can interact with, such as NATURAL PLAN, we include successful plans into the prompt for fewshot learning.
    
\textbf{ReAct} \cite{yao2023react}: Interleaves reasoning traces and task-specific actions, enabling LLM agents to act and reflect simultaneously.
        
\textbf{Reflexion} \cite{shinn2023reflexion}: Generates verbal reflections based on task feedback and stores them in an episodic memory buffer to guide future decision-making.
        
\textbf{Expel} \cite{expel}: Automatically gathers experience from a set of training tasks and extracts insights from successful trajectories or paired successful and failed trajectories.
        
\textbf{Inter-task Error Reflection}: Uses static, generated error-type reflections obtained from Inter-Task Learning to guide agent behavior.

\textbf{Retroformer Variant (DPO replacement)} \cite{yao2024retroformer}: Retroformer learns a retrospective model that tunes language agent prompts based on environment feedback via supervised fine-tuning (SFT) and Proximal Policy Optimization (PPO). However, tasks such as TravelPlanner often involve input lengths exceeding 10k tokens due to long trajectories and rich background context, making PPO-based training memory-intensive. To address this, we replace PPO with Direct Preference Optimization (DPO) for better efficiency. Our comparison on HotpotQA shows that this DPO-based variant achieves performance comparable to the original Retroformer (see details in Appendix $\S$ \ref{sec:hotpotqa}).

\subsection{Evaluation Metrics}

For TravelPlanner and Tau-bench, we adopt the official evaluation scripts to compute the \textbf{Pass Rate}. For NATURAL PLAN, each instance is annotated with a single ground-truth plan. We follow the original work to use \textbf{EM-based Accuracy}: accuracy is calculated as the proportion of output plans that exactly match the annotated reference plan.

\begin{table*}[ht]
\centering
\scalebox{0.8}{
\begin{tabular}{lc|lcc}
\toprule
  \multicolumn{2}{c}{\bf TravelPlanner} & \multicolumn{3}{c}{\bf NATURAL PLAN} \\ Methods &  Pass Rate &  Methods & Accuracy (Trip) & Accuracy (Meeting) \\ \midrule 
 ReAct & 4.44 & Fewshot & 44.06 & 38.50  \\
 Reflexion & 5.56 & Reflexion & 50.00 & 40.50  \\
 Inter-Task Error Reflection & 9.44 & Inter-Task Error Reflection & 51.56 & 42.00  \\
 Expel & 0.00 & Expel & 53.79 & 41.50  \\
 Retroformer variant & 12.78 & Retroformer variant & 47.50 & 44.00  \\
 \mymethod & \bf 20.00 & \mymethod & \bf 60.31 & \bf 48.50  \\
\bottomrule

\end{tabular}}
\caption{\label{main_result} Experimental results on TravelPlanner and NATURAL PLAN with Claude 3.5 Sonnet as the actor model.}
\vspace{-0.2cm}
\end{table*}

\begin{figure*}[ht]
\centering
\includegraphics[width=\linewidth]{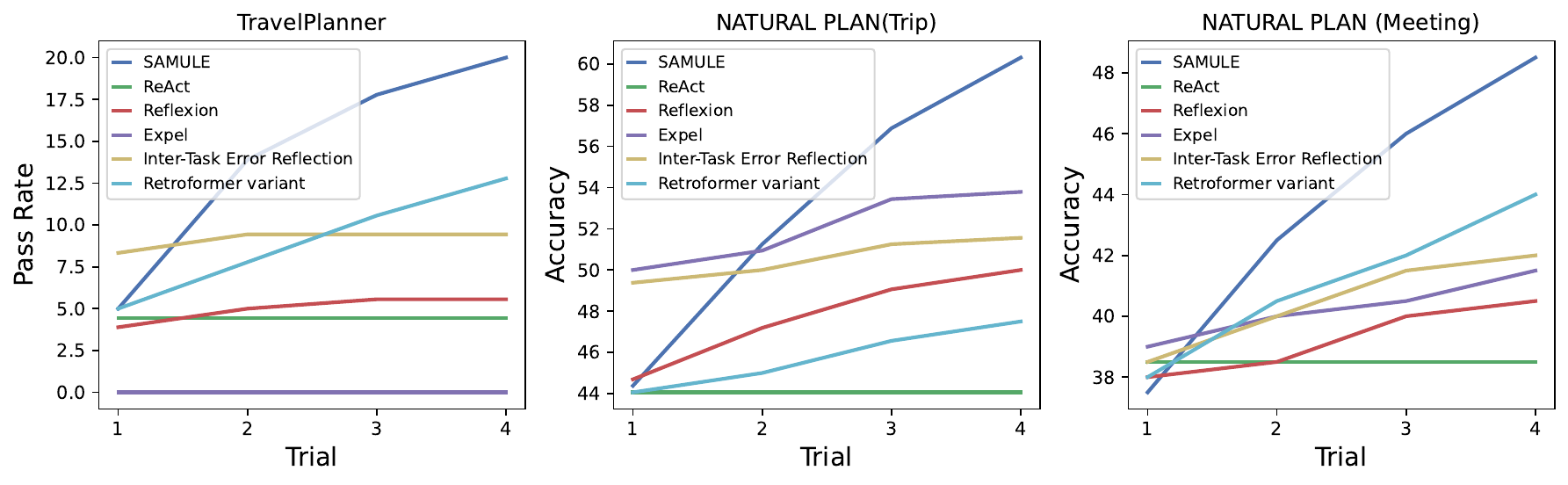}
    \caption{The performance change of different approaches across trials.} 
    \label{fig:change}
\vspace{-0.4cm}
\end{figure*}

\subsection{Main Results}
We first present results and the performance change across trial numbers on TravelPlanner and NATURAL PLAN with Claude 3.5 Sonnet-v2 as the actor model in Table~\ref{main_result} and Figure \ref{fig:change}. Across both benchmarks, reflection-based methods significantly outperform non-reflective baselines. Even the simplest reflective method, Reflexion, consistently improves over non-reflective methods, underscoring the value of incorporating reflection. Besides, we can also observe the following key findings:

\textbf{Cross-Trajectory Reflection Yields Gains.} Reflection across trajectories—such as Inter-task Error Reflection and our approach—substantially outperform single trajectory reflection (Reflexion). On TravelPlanner, Inter-task reaches 9.44\%, compared to Reflexion's 5.56\%, while our approach achieves 20\%. A similar trend is observed on NATURAL PLAN, such as with our method at 60.31\%, Inter-Task Error Reflection at 51.56\% versus 50\% (Reflexion) on the Trip domain. These results suggest that \textbf{structured, cross-trial reflection leads to more robust improvement} by abstracting recurring failure patterns across diverse task queries.

\textbf{Training Methods Comparison.} Comparing our approach to Inter-task Error Reflection reveals the benefits from training a retrospective language model via SFT. While both methods leverage cross-trial reflection, our model is fine-tuned to dynamically generate trajectory-specific reflections. This leads to improvements from 9.44\% to 20\% on TravelPlanner, from 51.56\% to 60.31\% on NATURAL PLAN Trip and from 42\% to 48.5\% on NATURAL PLAN Meeting. In contrast, Inter-task relies on static, precomputed reflections, which may include irrelevant errors unrelated to the current context. 

Moreover, although the Retroformer variant adopts more sophisticated RL techniques, it underperforms relative to our simpler SFT. This indicates that high-quality reflection synthesis is critical for training effective retrospective models; even advanced RL methods struggle if the reflection data is uninformative. These findings highlight that \textbf{well-designed reflection synthesis, even combined with simple SFT, can yield superior results while being more computationally efficient}.

\textbf{Reflexion Struggles in Complex Tasks.} Reflexion and its derivatives—Expel and Retroformer variant—struggle on more challenging benchmarks. In TravelPlanner, Reflexion achieves 5.56\%, Retroformer variant improves modestly to 12.78\%, and Expel collapses to 0\%. These results suggest that \textbf{self-reflection alone is insufficient when agents lack a clear understanding of their failure modes}. While Expel and Retroformer variant extend Reflexion with additional prompts or fine-tuning, they remain ineffective under high task complexity. In contrast, our method focuses on diverse error types derived from multiple failures and involve reference to facilitate error analysis, enabling more informative and actionable reflection even in complex, low-success-rate environments.

\textbf{Failure Provides Stronger Signals.} Expel performs well on NATURAL PLAN (53.79\% for Trip and 41.5\% for Meeting) but fails on TravelPlanner (0\%), likely due to its reliance on successful and paired (success/failure) trajectories, which are rare in harder tasks. These results reveal the \textbf{limitations of success-based reflection in high-error domains} and we show the generated insights by Expel and the detailed analysis in Appendix $\S$\ref{sec:expel}. Our approach instead focuses on failure, using error classification and clustering to extract insight from unsuccessful trials. This strategy proves more effective in environments where success is scarce but failures are abundant and informative by comparing with references. It underscores the value of failure-driven learning as a resilient strategy across tasks of varying complexity.

To further demonstrate the generality and effectiveness of our approach, we evaluated \mymethod using Claude 3.7 Sonnet as the actor on TravelPlanner in Table ~\ref{main_result_3.7}. The promising results confirm that our approach consistently improves agent performance across different actor models.

\begin{table}[t]
\centering
\scalebox{0.8}{
\begin{tabular}{lc}
\toprule
   \bf Methods &  \bf Pass Rate  \\ \midrule 
 ReAct & 9.44 \\
 Reflexion & 16.67 \\
 Inter-Task Error Reflection & 18.89 \\
 Expel & 17.22 \\
 Retroformer variant & 21.67 \\
 \mymethod & \bf 29.44\\
\bottomrule

\end{tabular}}
\caption{\label{main_result_3.7} Experimental results on TravelPlanner with Claude 3.7 Sonnet as the actor model.}
\vspace{-0.5cm}
\end{table}

\begin{table*}[ht]
\centering
\scalebox{0.8}{
\begin{tabular}{lcc|cc}
\toprule
  \multicolumn{1}{c}{\multirow{2}{*}{\bf Methods}} & \multicolumn{2}{c}{\bf Non-Interactive (3 trials)} & \multicolumn{2}{c}{\bf Interactive} \\ &  Pass Rate (Retail) &  Pass Rate (Airline) &  Pass Rate (Retail) &  Pass Rate (Airline) \\ \midrule 
 ReAct & 64.87 & 46.40 & 64.87 & 46.40  \\
 Reflexion & 82.61 & 56.00 & 69.75 & 48.50  \\
 \mymethod & \bf 87.83 & \bf 66.00 & \bf 75.97 & \bf 55.32  \\
\bottomrule
\end{tabular}}
\caption{\label{tau_result} Experimental results on Tau-Bench.}
\vspace{-0.5cm}
\end{table*}

\subsection{Interactive vs. Non-Interactive}

We evaluate our framework under both non-interactive and interactive settings using Tau-Bench, and show the results in Table~\ref{tau_result}.

In the \textbf{non-interactive} setting—consistent with the previous experiments on TravelPlanner and NATURAL PLAN—our method consistently outperforms baselines in both domains. SEMR achieves 87.83\% and 66\% pass rates in Retail and Airline, respectively, surpassing Reflexion (82.61\%, 56\%) and ReAct (64.87\%, 46.4\%). These results further confirm the effectiveness of our multi-level reflection mechanism when applied to complete trajectories.

In the \textbf{interactive} setting, our approach also outperforms baselines, achieving 75.97\% and 55.32\% pass rates in Retail and Airline, respectively. Notably, Reflexion also benefits from interaction, improving over its non-interactive performance in Retail. This suggests that real-time feedback allows reflective agents to detect and correct partial trajectory errors, leading to better outcomes.

Overall, these results demonstrate that our framework generalizes well across both interactive and non-interactive contexts. The ability to reflect both retrospectively (on complete trajectories) and incrementally (during interactions) makes our method a practical solution for real-world applications requiring adaptive and context-aware reasoning.

\subsection{Ablation Study: When to Add Reference?}

A key constraint of our proposed framework is its reliance on a reference output during training, such as $y_\text{train}^{(i)}$, which is sometimes unavailable in many existing benchmarks. We hypothesize that such references provide critical guidance for error analysis—especially in complex tasks where agents frequently fail and self-reflection alone may lead to hallucination or superficial feedback.

We therefore investigate the impact of providing references at different stages of our multi-level reflection synthesis. We focus on the TravelPlanner dataset and compare the following three variants:
\begin{itemize}
    \item \textbf{No Reference}: No reference $y_\text{train}^{(i)}$is provided.
    \item \textbf{Single + Intra (Reference)}: The reference $y_\text{train}^{(i)}$ is provided during both Single Trajectory and Intra-Task Learning stages.
    \item \textbf{Single (Reference) + Intra (No Reference)}: The reference $y_\text{train}^{(i)}$ is provided only during the Single Trajectory Learning, but omitted during Intra-Task Learning.
\end{itemize}

\begin{table}[t]
\small
\centering
\scalebox{0.9}{
\begin{tabular}{p{5.2cm}c}
\toprule
\bf Method Variant & \bf Pass Rate (\%) \\ \midrule 
No Reference & 18.33 \\
Single + Intra (Reference) & 15.56 \\
Single (Reference) + Intra (No Reference) & \textbf{20.00} \\
\bottomrule
\end{tabular}}
\caption{\label{ref_results} Pass rates on TravelPlanner for different strategies of providing reference during reflection.}
\vspace{-0.4cm}
\end{table}

Table~\ref{ref_results} shows that providing reference only during the Single Trajectory Learning stage yields the best performance (20.00\%), outperforming both the no-reference variant (18.33\%) and the variant that includes references in both stages (15.56\%). This supports our hypothesis that \textbf{reference can be particularly useful at the micro level}, where the model benefits from detailed, item-by-item comparison between its trajectory and the reference. For example, if the agent selects a restaurant not present in the reference plan, it can reflect on the discrepancy and infer a potential mistake, such as choosing a restaurant scheduled before arrival time.

Surprisingly, providing the reference at both Single Trajectory and Intra-Task Learning  results in degraded performance. We attribute this to the benchmarks like TravelPlanner, where $y_\text{train}^{(i)}$ is not a unique solution. \textbf{Overexposing the reference may narrow the model’s focus}, encouraging it to align too closely with one specific plan and ignore other legitimate errors not captured by the reference. This suggests that while micro-level references enrich error reasoning, excessive reliance on them during the meso-level can reduce diversity in error detection and hinder generalization. 

\begin{table*}[t]
\centering
\scalebox{0.8}{
\begin{tabular}{lccc}
\toprule
   \bf Methods &  \bf TravelPlanner & \bf NATURAL PLAN (Trip) & \bf NATURAL PLAN (Meeting) \\ \midrule 
 Reflexion & 0.13 & 0.42 & 0.22 \\
 \mymethod & \bf 0.67 & \bf 0.73 & \bf 0.53 \\
\bottomrule

\end{tabular}}
\caption{\label{error_reduction} Error reduction rates achieved by our synthesized reflections and Reflexion.}
\vspace{-0.5cm}
\end{table*}

\subsection{Error Reduction Analysis}
One key feature of \mymethod is that we focus on addressing errors, especially we perform error taxonomy construction and error classification during the meso level. To further validate the effectiveness of our reflections in addressing these errors, we conducted an error reduction evaluation by adding final synthesized reflections to agents, retrying queries, and reclassifying the resulting errors. We compare it with the reflections generated by Reflexion and show the results in Table \ref{error_reduction}. This quantitative result shows that our proposed multi-level and error-prone reflection synthesis significantly improves the agent’s ability to address identified error types.

\subsection{Qualitative Analysis}
Due to limited space, we provide some qualitative examples, including an example of constructed error taxonomy during the meso-level of our proposed Multi-level Reflection Synthesis in Appendix $\S$ \ref{sec:error_taxonomy} and a comparison between our approach and the Retroformer Variant on TravelPlanner in Appendix $\S$\ref{sec:qualitative}. The qualitative comparison illustrates that our trained retrospective model identifies the true issues in the agent’s output plan, enabling the agent to correct its mistakes in the subsequent trial. In contrast, the Retroformer Variant incorrectly diagnoses irrelevant errors, which biases the agent toward focusing on unrelated factors—such as geographic constraints and meal timing considerations—that are not actually implicated in the original plan. This highlights the superior diagnostic capability of our approach in guiding more targeted and meaningful corrections.

\section{Conclusion}

In this work, we introduced \mymethod, a new framework for self-learning LLM agents via training a retrospective language model enhanced by multi-level reflection synthesis. By focusing on the Single-Trajectory, Intra-Task, and Inter-Task learning, our method synthesizes reflections via systematically analyzing past trajectories from micro-levl to macro-level, which identifies recurring error patterns and formulates more effective plans. To operationalize this reflection-driven learning, we train a retrospective language model using the synthesized reflections. Experimental results on the TravelPlanner, NATURAL PLAN, and Tau-bench demonstrate that our approach improves agent performance in complex planning tasks, underscoring the value of structured, cross-trial reflection for self-improvement. Our findings highlight the potential of multi-level reflective learning via SFT as a general paradigm for enhancing the performance of LLM agents.

\section{Limitations}

While our method demonstrates strong performance and significantly enhances agents' ability to learn from failures through multi-level reflection, several limitations remain:

\textbf{Static Error Taxonomy Limits Continual Learning.}  
Our framework constructs an error taxonomy during the offline reflection synthesis process to guide failure analysis and reflection generation. However, this taxonomy remains static throughout inference. As agents encounter new tasks or previously unseen failure patterns, the existing taxonomy may become incomplete or outdated. Future work could explore incremental taxonomy construction and online adaptation methods to support lifelong learning in dynamic environments.

\textbf{Computational Overhead in Multi-Level Reflection Synthesis.}  
Generating and organizing reflection data at multiple levels introduces additional computational costs during the data preparation phase. Although the final retrospective model is lightweight and efficient at inference time, the offline processes of trajectory analysis, error taxonomy construction, and cross-task clustering are resource-intensive, especially for large-scale datasets with long trajectories. Future research could investigate more scalable reflection synthesis techniques or efficient memory management strategies to mitigate this overhead.

\bibliography{custom}

\begin{thebibliography}{45}
\providecommand{\natexlab}[1]{#1}

\bibitem[{Alabdulkarim et~al.(2022)Alabdulkarim, Singh, Mansi, Hall, and Riedl}]{experiential_explanations_2022_2210}
Amal Alabdulkarim, Madhuri Singh, Gennie Mansi, Kaely Hall, and Mark~O. Riedl. 2022.
\newblock \href {https://arxiv.org/abs/2210.04723v4} {Experiential explanations for reinforcement learning}.
\newblock \emph{arXiv preprint}.

\bibitem[{Chen et~al.(2025)Chen, Chen, Sun, Liu, and Gan}]{scaling_autonomous_2025_2502}
Zhenfang Chen, Delin Chen, Rui Sun, Wenjun Liu, and Chuang Gan. 2025.
\newblock \href {https://arxiv.org/abs/2502.12130v1} {Scaling autonomous agents via automatic reward modeling and planning}.
\newblock \emph{arXiv preprint}.

\bibitem[{Feng et~al.(2025)Feng, Lan, Dai, Wang, Tang, Chen, Dong, and Wen}]{improving_retrospective_2025_2503}
Xueyang Feng, Bo~Lan, Quanyu Dai, Lei Wang, Jiakai Tang, Xu~Chen, Zhenhua Dong, and Ji-Rong Wen. 2025.
\newblock \href {https://arxiv.org/abs/2503.01490v1} {Improving retrospective language agents via joint policy gradient optimization}.
\newblock \emph{arXiv preprint}.

\bibitem[{Fourney et~al.(2024)Fourney, Bansal, Mozannar, Tan, Salinas, Niedtner, Proebsting, Bassman, Gerrits, Alber et~al.}]{fourney2024magentic}
Adam Fourney, Gagan Bansal, Hussein Mozannar, Cheng Tan, Eduardo Salinas, Friederike Niedtner, Grace Proebsting, Griffin Bassman, Jack Gerrits, Jacob Alber, and 1 others. 2024.
\newblock Magentic-one: A generalist multi-agent system for solving complex tasks.
\newblock \emph{arXiv preprint arXiv:2411.04468}.

\bibitem[{Gao et~al.(2024)Gao, Ding, Cui, Zhao, Wang, Liu, and Qin}]{gao-etal-2024-self-evolving}
Jinglong Gao, Xiao Ding, Yiming Cui, Jianbai Zhao, Hepeng Wang, Ting Liu, and Bing Qin. 2024.
\newblock \href {https://doi.org/10.18653/v1/2024.acl-long.346} {Self-evolving {GPT}: A lifelong autonomous experiential learner}.
\newblock In \emph{Proceedings of the 62nd Annual Meeting of the Association for Computational Linguistics (Volume 1: Long Papers)}, pages 6385--6432, Bangkok, Thailand. Association for Computational Linguistics.

\bibitem[{Ge et~al.(2021)Ge, Dinh, Liu, Su, Lu, Wang, and Diesner}]{ge2021baco}
Yubin Ge, Ly~Dinh, Xiaofeng Liu, Jinsong Su, Ziyao Lu, Ante Wang, and Jana Diesner. 2021.
\newblock Baco: A background knowledge-and content-based framework for citing sentence generation.
\newblock In \emph{Proceedings of the 59th Annual Meeting of the Association for Computational Linguistics and the 11th International Joint Conference on Natural Language Processing (Volume 1: Long Papers)}, pages 1466--1478.

\bibitem[{Ge et~al.(2023)Ge, Hazarika, Liu, and Namazifar}]{ge2023supervised}
Yubin Ge, Devamanyu Hazarika, Yang Liu, and Mahdi Namazifar. 2023.
\newblock Supervised fine-tuning of large language models on human demonstrations through the lens of memorization.
\newblock In \emph{NeurIPS 2023 Workshop on Instruction Tuning and Instruction Following}.

\bibitem[{Ge et~al.(2025)Ge, Romeo, Cai, Shu, Sunkara, Benajiba, and Zhang}]{ge2025tremu}
Yubin Ge, Salvatore Romeo, Jason Cai, Raphael Shu, Monica Sunkara, Yassine Benajiba, and Yi~Zhang. 2025.
\newblock Tremu: Towards neuro-symbolic temporal reasoning for llm-agents with memory in multi-session dialogues.
\newblock \emph{arXiv preprint arXiv:2502.01630}.

\bibitem[{Golchha et~al.(2024)Golchha, Yerawar, Patel, Dan, and Murugesan}]{golchha-etal-2024-language}
Hitesh Golchha, Sahil Yerawar, Dhruvesh Patel, Soham Dan, and Keerthiram Murugesan. 2024.
\newblock \href {https://doi.org/10.18653/v1/2024.findings-naacl.7} {Language guided exploration for {RL} agents in text environments}.
\newblock In \emph{Findings of the Association for Computational Linguistics: NAACL 2024}, pages 93--102, Mexico City, Mexico. Association for Computational Linguistics.

\bibitem[{Gupta et~al.(2024)Gupta, Kirtania, Singha, Gulwani, Radhakrishna, Soares, and Shi}]{gupta-etal-2024-metareflection}
Priyanshu Gupta, Shashank Kirtania, Ananya Singha, Sumit Gulwani, Arjun Radhakrishna, Gustavo Soares, and Sherry Shi. 2024.
\newblock \href {https://doi.org/10.18653/v1/2024.emnlp-main.477} {{M}eta{R}eflection: Learning instructions for language agents using past reflections}.
\newblock In \emph{Proceedings of the 2024 Conference on Empirical Methods in Natural Language Processing}, pages 8369--8385, Miami, Florida, USA. Association for Computational Linguistics.

\bibitem[{Hu et~al.(2022)Hu, Wallis, Allen-Zhu, Li, Wang, Wang, Chen et~al.}]{hulora}
Edward~J Hu, Phillip Wallis, Zeyuan Allen-Zhu, Yuanzhi Li, Shean Wang, Lu~Wang, Weizhu Chen, and 1 others. 2022.
\newblock Lora: Low-rank adaptation of large language models.
\newblock In \emph{International Conference on Learning Representations}.

\bibitem[{Ji et~al.(2024)Ji, Wu, Ma, Li, and Wang}]{ji2024testing}
Zhenlan Ji, Daoyuan Wu, Pingchuan Ma, Zongjie Li, and Shuai Wang. 2024.
\newblock Testing and understanding erroneous planning in llm agents through synthesized user inputs.
\newblock \emph{arXiv preprint arXiv:2404.17833}.

\bibitem[{Lee et~al.(2025)Lee, Fischer, Wu, Marwood, Baluja, Schuurmans, and Chen}]{lee2025evolving}
Kuang-Huei Lee, Ian Fischer, Yueh-Hua Wu, Dave Marwood, Shumeet Baluja, Dale Schuurmans, and Xinyun Chen. 2025.
\newblock Evolving deeper llm thinking.
\newblock \emph{arXiv preprint arXiv:2501.09891}.

\bibitem[{Li et~al.(2024)Li, Yang, and Ettinger}]{li2024hindsight}
Yanhong Li, Chenghao Yang, and Allyson Ettinger. 2024.
\newblock When hindsight is not 20/20: Testing limits on reflective thinking in large language models.
\newblock In \emph{Findings of the Association for Computational Linguistics: NAACL 2024}, pages 3741--3753.

\bibitem[{Liu et~al.(2024)Liu, AlDahoul, Eady, Zaki, and Rahwan}]{selfreflection_makes_2024}
Fengyuan Liu, Nouar AlDahoul, Gregory Eady, Yasir Zaki, and Talal Rahwan. 2024.
\newblock \href {https://arxiv.org/abs/2406.10400v2} {Self-reflection makes large language models safer, less biased, and ideologically neutral}.
\newblock \emph{arXiv preprint}.

\bibitem[{Loshchilov and Hutter(2018)}]{loshchilov2018decoupled}
Ilya Loshchilov and Frank Hutter. 2018.
\newblock Decoupled weight decay regularization.
\newblock In \emph{International Conference on Learning Representations (ICLR)}.

\bibitem[{McLeod(2017)}]{mcleod2017kolb}
Saul McLeod. 2017.
\newblock Kolb's learning styles and experiential learning cycle.
\newblock \emph{Simply psychology}, 5.

\bibitem[{Qi et~al.(2024)Qi, Liu, Iong, Lai, Sun, Zhao, Yang, Yang, Sun, Yao, Zhang, Xu, Tang, and Dong}]{webrl_training_2024_2411}
Zehan Qi, Xiao Liu, Iat~Long Iong, Hanyu Lai, Xueqiao Sun, Wenyi Zhao, Yu~Yang, Xinyue Yang, Jiadai Sun, Shuntian Yao, Tianjie Zhang, Wei Xu, Jie Tang, and Yuxiao Dong. 2024.
\newblock \href {https://arxiv.org/abs/2411.02337v3} {Webrl: Training llm web agents via self-evolving online curriculum reinforcement learning}.
\newblock \emph{arXiv preprint}.

\bibitem[{Qian et~al.(2024)Qian, Dang, Li, Liu, Xie, Wang, Chen, Yang, Cong, Che, Liu, and Sun}]{qian-etal-2024-experiential}
Chen Qian, Yufan Dang, Jiahao Li, Wei Liu, Zihao Xie, YiFei Wang, Weize Chen, Cheng Yang, Xin Cong, Xiaoyin Che, Zhiyuan Liu, and Maosong Sun. 2024.
\newblock \href {https://doi.org/10.18653/v1/2024.acl-long.305} {Experiential co-learning of software-developing agents}.
\newblock In \emph{Proceedings of the 62nd Annual Meeting of the Association for Computational Linguistics (Volume 1: Long Papers)}, pages 5628--5640, Bangkok, Thailand. Association for Computational Linguistics.

\bibitem[{Radha et~al.(2024)Radha, Jelyani, Ghukasyan, and Goktas}]{iteration_of_2024_2409}
Santosh~Kumar Radha, Yasamin~Nouri Jelyani, Ara Ghukasyan, and Oktay Goktas. 2024.
\newblock \href {https://arxiv.org/abs/2409.12618v2} {Iteration of thought: Leveraging inner dialogue for autonomous large language model reasoning}.
\newblock \emph{arXiv preprint}.

\bibitem[{Renze and Guven(2024)}]{selfreflection_in_2024_2405}
Matthew Renze and Erhan Guven. 2024.
\newblock \href {https://arxiv.org/abs/2405.06682v3} {Self-reflection in llm agents: Effects on problem-solving performance}.
\newblock \emph{arXiv preprint}.

\bibitem[{Shen et~al.(2025)Shen, Shu, Pratik, Gung, Ge, Sunkara, and Zhang}]{shen2025optimizing}
Ming Shen, Raphael Shu, Anurag Pratik, James Gung, Yubin Ge, Monica Sunkara, and Yi~Zhang. 2025.
\newblock Optimizing llm-based multi-agent system with textual feedback: A case study on software development.
\newblock \emph{arXiv preprint arXiv:2505.16086}.

\bibitem[{Shinn et~al.(2023)Shinn, Cassano, Gopinath, Narasimhan, and Yao}]{shinn2023reflexion}
Noah Shinn, Federico Cassano, Ashwin Gopinath, Karthik Narasimhan, and Shunyu Yao. 2023.
\newblock Reflexion: Language agents with verbal reinforcement learning.
\newblock \emph{Advances in Neural Information Processing Systems}, 36:8634--8652.

\bibitem[{Sun et~al.(2023)Sun, Yang, Jiralerspong, Malenfant, Alsbury-Nealy, Bengio, and Richards}]{sun2023contrastive}
Chen Sun, Wannan Yang, Thomas Jiralerspong, Dane Malenfant, Benjamin Alsbury-Nealy, Yoshua Bengio, and Blake Richards. 2023.
\newblock Contrastive retrospection: honing in on critical steps for rapid learning and generalization in rl.
\newblock \emph{Advances in Neural Information Processing Systems}, 36:31117--31139.

\bibitem[{Tao et~al.(2024)Tao, Lin, Chen, Li, Wu, Li, Jin, Huang, Tao, and Zhou}]{a_survey_2024_2404}
Zhengwei Tao, Ting-En Lin, Xiancai Chen, Hangyu Li, Yuchuan Wu, Yongbin Li, Zhi Jin, Fei Huang, Dacheng Tao, and Jingren Zhou. 2024.
\newblock \href {https://arxiv.org/abs/2404.14387v2} {A survey on self-evolution of large language models}.
\newblock \emph{arXiv preprint}.

\bibitem[{Tu et~al.(2024)Tu, Sun, Zhang, Lan, and Zhao}]{online_preferencebased_2024_2412}
Songjun Tu, Jingbo Sun, Qichao Zhang, Xiangyuan Lan, and Dongbin Zhao. 2024.
\newblock \href {https://arxiv.org/abs/2412.16878v1} {Online preference-based reinforcement learning with self-augmented feedback from large language model}.
\newblock \emph{arXiv preprint}.

\bibitem[{Tulis et~al.(2016)Tulis, Steuer, and Dresel}]{tulis2016learning}
Maria Tulis, Gabriele Steuer, and Markus Dresel. 2016.
\newblock Learning from errors: a model of individual processes.
\newblock \emph{Frontline Learning Research}, 4(2):12--26.

\bibitem[{Wang et~al.(2022)Wang, Li, Li, Xia, Wang, Xie, and Wu}]{wang2022learning}
Chengwei Wang, Junyi Li, Haiyan Li, Yijing Xia, Xiaoyu Wang, Yufei Xie, and Jinyang Wu. 2022.
\newblock Learning from errors? the impact of erroneous example elaboration on learning outcomes of medical statistics in chinese medical students.
\newblock \emph{BMC medical education}, 22(1):469.

\bibitem[{Wang et~al.(2024{\natexlab{a}})Wang, Wang, Su, Tong, and Song}]{self_ev}
Qineng Wang, Zihao Wang, Ying Su, Hanghang Tong, and Yangqiu Song. 2024{\natexlab{a}}.
\newblock \href {https://arxiv.org/abs/2402.18272} {Rethinking the bounds of llm reasoning: Are multi-agent discussions the key?}
\newblock \emph{Preprint}, arXiv:2402.18272.

\bibitem[{Wang et~al.(2024{\natexlab{b}})Wang, Wang, Su, Tong, and Song}]{adapt}
Qineng Wang, Zihao Wang, Ying Su, Hanghang Tong, and Yangqiu Song. 2024{\natexlab{b}}.
\newblock \href {https://arxiv.org/abs/2402.18272} {Rethinking the bounds of llm reasoning: Are multi-agent discussions the key?}
\newblock \emph{Preprint}, arXiv:2402.18272.

\bibitem[{Wang et~al.(2024{\natexlab{c}})Wang, Li, Han, Zhang, and Baldwin}]{wang2024learning}
Renxi Wang, Haonan Li, Xudong Han, Yixuan Zhang, and Timothy Baldwin. 2024{\natexlab{c}}.
\newblock Learning from failure: Integrating negative examples when fine-tuning large language models as agents.
\newblock \emph{arXiv preprint arXiv:2402.11651}.

\bibitem[{Wei et~al.(2022)Wei, Wang, Schuurmans, Bosma, Xia, Chi, Le, Zhou et~al.}]{wei2022chain}
Jason Wei, Xuezhi Wang, Dale Schuurmans, Maarten Bosma, Fei Xia, Ed~Chi, Quoc~V Le, Denny Zhou, and 1 others. 2022.
\newblock Chain-of-thought prompting elicits reasoning in large language models.
\newblock \emph{Advances in neural information processing systems}, 35:24824--24837.

\bibitem[{Xie et~al.(2024)Xie, Zhang, Chen, Zhu, Lou, Tian, Xiao, and Su}]{xie2024travelplanner}
Jian Xie, Kai Zhang, Jiangjie Chen, Tinghui Zhu, Renze Lou, Yuandong Tian, Yanghua Xiao, and Yu~Su. 2024.
\newblock Travelplanner: a benchmark for real-world planning with language agents.
\newblock In \emph{Proceedings of the 41st International Conference on Machine Learning}, pages 54590--54613.

\bibitem[{Xie et~al.(2025)Xie, Chen, Mao, Xu, Kong et~al.}]{xie2025teaching}
Zhihui Xie, Liyu Chen, Weichao Mao, Jingjing Xu, Lingpeng Kong, and 1 others. 2025.
\newblock Teaching language models to critique via reinforcement learning.
\newblock \emph{arXiv preprint arXiv:2502.03492}.

\bibitem[{Xiong et~al.(2025)Xiong, Zhang, Ye, Chen, Jiang, and Zhang}]{selfrewarding_correction_2025_2502}
Wei Xiong, Hanning Zhang, Chenlu Ye, Lichang Chen, Nan Jiang, and Tong Zhang. 2025.
\newblock \href {https://arxiv.org/abs/2502.19613v1} {Self-rewarding correction for mathematical reasoning}.
\newblock \emph{arXiv preprint}.

\bibitem[{Yang et~al.(2018)Yang, Qi, Zhang, Bengio, Cohen, Salakhutdinov, and Manning}]{yang2018hotpotqa}
Zhilin Yang, Peng Qi, Saizheng Zhang, Yoshua Bengio, William Cohen, Ruslan Salakhutdinov, and Christopher~D Manning. 2018.
\newblock Hotpotqa: A dataset for diverse, explainable multi-hop question answering.
\newblock In \emph{Proceedings of the 2018 Conference on Empirical Methods in Natural Language Processing}, pages 2369--2380.

\bibitem[{Yao et~al.(2024{\natexlab{a}})Yao, Shinn, Razavi, and Narasimhan}]{yaotau}
Shunyu Yao, Noah Shinn, Pedram Razavi, and Karthik~R Narasimhan. 2024{\natexlab{a}}.
\newblock $\tau$-bench: A benchmark for tool-agent-user interaction in real-world domains.
\newblock In \emph{The Thirteenth International Conference on Learning Representations}.

\bibitem[{Yao et~al.(2023)Yao, Zhao, Yu, Du, Shafran, Narasimhan, and Cao}]{yao2023react}
Shunyu Yao, Jeffrey Zhao, Dian Yu, Nan Du, Izhak Shafran, Karthik Narasimhan, and Yuan Cao. 2023.
\newblock React: Synergizing reasoning and acting in language models.
\newblock In \emph{International Conference on Learning Representations (ICLR)}.

\bibitem[{Yao et~al.(2024{\natexlab{b}})Yao, Heinecke, Niebles, Liu, Feng, Xue, N, Chen, Zhang, Arpit, Xu, Mui, Wang, Xiong, and Savarese}]{yao2024retroformer}
Weiran Yao, Shelby Heinecke, Juan~Carlos Niebles, Zhiwei Liu, Yihao Feng, Le~Xue, Rithesh~R N, Zeyuan Chen, Jianguo Zhang, Devansh Arpit, Ran Xu, Phil~L Mui, Huan Wang, Caiming Xiong, and Silvio Savarese. 2024{\natexlab{b}}.
\newblock \href {https://openreview.net/forum?id=KOZu91CzbK} {Retroformer: Retrospective large language agents with policy gradient optimization}.
\newblock In \emph{The Twelfth International Conference on Learning Representations}.

\bibitem[{Yin et~al.(2024)Yin, Wang, Pan, Wan, and Wang}]{gödel_agent_2024_2410}
Xunjian Yin, Xinyi Wang, Liangming Pan, Xiaojun Wan, and William~Yang Wang. 2024.
\newblock \href {https://arxiv.org/abs/2410.04444v3} {Gödel agent: A self-referential agent framework for recursive self-improvement}.
\newblock \emph{arXiv preprint}.

\bibitem[{Zhang et~al.(2024)Zhang, Shen, Wu, Peng, Wang, Zhuang, and Lu}]{selfcontrast_better_2024_2401}
Wenqi Zhang, Yongliang Shen, Linjuan Wu, Qiuying Peng, Jun Wang, Yueting Zhuang, and Weiming Lu. 2024.
\newblock \href {https://arxiv.org/abs/2401.02009v3} {Self-contrast: Better reflection through inconsistent solving perspectives}.
\newblock \emph{arXiv preprint}.

\bibitem[{Zhao et~al.(2024{\natexlab{a}})Zhao, Huang, Xu, Lin, Liu, and Huang}]{expel}
Andrew Zhao, Daniel Huang, Quentin Xu, Matthieu Lin, Yong-Jin Liu, and Gao Huang. 2024{\natexlab{a}}.
\newblock Expel: Llm agents are experiential learners.
\newblock In \emph{Proceedings of the AAAI Conference on Artificial Intelligence}, pages 19632--19642.

\bibitem[{Zhao et~al.(2024{\natexlab{b}})Zhao, Huang, Xu, Lin, Liu, and Huang}]{zhao2024expel}
Andrew Zhao, Daniel Huang, Quentin Xu, Matthieu Lin, Yong-Jin Liu, and Gao Huang. 2024{\natexlab{b}}.
\newblock Expel: Llm agents are experiential learners.
\newblock In \emph{Proceedings of the AAAI Conference on Artificial Intelligence}, volume~38, pages 19632--19642.

\bibitem[{Zheng et~al.(2024)Zheng, Mishra, Zhang, Chen, Chen, Nova, Hou, Cheng, Le, Chi et~al.}]{zheng2024natural}
Huaixiu~Steven Zheng, Swaroop Mishra, Hugh Zhang, Xinyun Chen, Minmin Chen, Azade Nova, Le~Hou, Heng-Tze Cheng, Quoc~V Le, Ed~H Chi, and 1 others. 2024.
\newblock Natural plan: Benchmarking llms on natural language planning.
\newblock \emph{arXiv preprint arXiv:2406.04520}.

\bibitem[{Zhou et~al.(2023)Zhou, Madaan, Potharaju, Gupta, McKee, Holtzman, Pujara, Ren, Mishra, Nematzadeh et~al.}]{zhou2023far}
Pei Zhou, Aman Madaan, Srividya~Pranavi Potharaju, Aditya Gupta, Kevin~R McKee, Ari Holtzman, Jay Pujara, Xiang Ren, Swaroop Mishra, Aida Nematzadeh, and 1 others. 2023.
\newblock How far are large language models from agents with theory-of-mind?
\newblock \emph{arXiv preprint arXiv:2310.03051}.

\end{thebibliography}
\nocite{*}

\appendix
\section{Implementation Details}
\label{sec:implementation}
We tested Claude 3.5 sonnet-v2 and Claude 3.7 sonnet as the backbone actor model for LLM agents, and also use Claude 3.5 sonnet-v2 to synthesize high-quality reflections through our multi-level reflection synthesis. For the non-interactive setting, we follow previous work \cite{yao2024retroformer, shinn2023reflexion} we run the agent for at most 4 trials.

As for training the retrospective model, we select QWEN 2.5 3B. Models are trained with LoRA \cite{hulora} and DeepSpeed stage 3 for saving memory, and are optimized using AdamW \citep{loshchilov2018decoupled}. We set the learning rate to $5e-5$, and the learning rate was updated using a linear decay schedule with an end value of 0. We set the total training epochs to $15$, and the batch size to $1$. The training was performed on $8$ NVIDIA A100 Tensor Core GPUs and it took about 1 hour for one training run. 

\section{Comparison of Retroformer and Its Variant on HotPotQA}  
\label{sec:hotpotqa}

While the original Retroformer~\cite{yao2024retroformer} has demonstrated effectiveness on relatively simple benchmarks such as HotPotQA~\cite{yang2018hotpotqa}, our research focuses on more complex and challenging agentic benchmarks where trajectories are significantly longer. For instance, on \textbf{TravelPlanner}, agent trajectories can exceed 10,000 tokens, introducing substantial challenges for model training due to long-context limitations and increased memory consumption.

To facilitate a fair comparison and mitigate out-of-memory issues caused by lengthy inputs, we evaluate the original Retroformer (SFT + reward model training + PPO) against its more memory-efficient variant (SFT + DPO) on the HotPotQA benchmark. Specifically, we follow the same reward shaping strategy used in Retroformer, where the reward is computed based on task improvement after incorporating a generated reflection. This allows us to pair high-reward samples with low-reward samples for PPO training. Both models are trained using the Qwen-2.5 3B language model.

\begin{table}[t]
\small
\centering
\scalebox{0.95}{
\begin{tabular}{p{4.7cm}c}
\toprule
\bf Method & \bf Success Rate (\%) \\ \midrule 
Retroformer (SFT + reward model training + PPO) & 43 \\
Retroformer Variant (SFT + DPO) & \bf 44 \\
\bottomrule
\end{tabular}}
\caption{\label{hotpotqa_results} Success rates of Retroformer and its variant on HotPotQA.}
\end{table}

As shown in Table~\ref{hotpotqa_results}, the Retroformer variant achieves a comparable success rate to the original Retroformer (44\% vs. 43\%). This result confirms our hypothesis that DPO can serve as a more memory-efficient alternative to PPO without sacrificing performance. Therefore, we adopt the Retroformer variant in our subsequent experiments to better accommodate the computational demands of long-context agentic tasks.

\section{Designed Prompts for Multi-Level Reflection Synthesis}
\label{sec:multi_prompt}
We list the designed prompts for our proposed multi-level reflection synthesis on TravelPlanner, and for other benchmarks we slightly modify the instruction to accommodate for the corresponding benchmark. Specifically, we show the prompt for Single-Level Learning in Figure \ref{fig:single_prompt}, the prompt for constructing the error taxonomy during Intra-Task Learning in Figure \ref{fig:error_taxonomy_prompt}, the prompt for error classification during Intra-Task Learning in Figure \ref{fig:error_classification_prompt} and the prompt for Inter-Task Learning in Figure \ref{fig:inter_prompt}.

\begin{figure*}[ht]
    \centering
    \begin{tcolorbox}[left=5pt,right=5pt,colback=white,colframe=black,boxrule=1pt,fontupper=\ttfamily]
        You are an advanced reasoning agent that can improve based on self refection. You will be given a previous reasoning trial in which you were given access to an automatic cost calculation environment, a travel query to give a plan and relevant information. Only the selection whose name and city match the given information will be calculated correctly. Meanwhile, you are also given one valid plan to the given query as a reference to facilitate your analysis, though not the only valid one. Now by comparing your previous trial with the valid plan and verifying factuality of your previous trial with the given information, give a comprehensive diagnosis of your previous trial in a few sentences without mentioning the valid plan. Then devise a new, concise, high level plan that aims to mitigate potential failures. Use complete sentences. \\

        Given information: \$\{text\} \\
        
        Previous trial: \\
        Query: \$\{query\} \$\{scratchpad\} \\
        
        Valid plan: \$\{annotated\_plan\} \\
        
        Reflection:
    \end{tcolorbox}
\caption{The designed prompt for Single-Level Learning.}
\label{fig:single_prompt}
\end{figure*}

\begin{figure*}[ht]
    \centering
    \begin{tcolorbox}[left=5pt,right=5pt,colback=white,colframe=black,boxrule=1pt,fontupper=\ttfamily]
    You are an advanced reasoning agent that aims at improving reasoning through self refection. You were tasked to give a valid plan to a travel query given its relevant information and an automatic cost calculation environment. Only the selection whose name and city match the given information will be calculated correctly. You have tried to solve this task multiple times, and you will also be given your previous reasoning paths of these trials with their final results (either success or fail). Now, you need to analyze all reasoning paths of these trials by checking whether each action step leads to a successful final plan and what types of error you have made at each action step. \\
    
    Step-by-step Instruction:\\
    1. Read all information and analyze all given trials based on the given information. Carefully check how failed trials violate information or constraints based on the meta data from the given information, which caused the failure. \\
    2. Based on your analysis, come up with a complete error taxonomy for categorizing the all common errors that you have made for actions in different failed trials. You should merge similar error types as one. You will be given the error taxonomy from previous reflections, and you can add new error types. \\
    3. Generate your output in JSON format as one dictionary which contains the following keys: \\ 
    \quad a. "error\_taxonomy": a list of strings containing all error type names without explanation. \\ 
    \quad b. "rationale": your explanation of giving this error taxonomy including how the failed trials fail and how the valid plan, and successful trials success. \\
    
    Given Information: \$\{text\} \\
    
    Query: \$ \{query\} \\
    
    Previous Trials:
    \$ \{trials\} \\

    Error Taxonomy:
    \$ \{error\_taxonomy\}
    \end{tcolorbox}
\caption{The designed prompt for constructing the error taxonomy during the Intra-Level Learning.}
\label{fig:error_taxonomy_prompt}
\end{figure*}

\begin{figure*}[ht]
    \centering
    \begin{tcolorbox}[left=5pt,right=5pt,colback=white,colframe=black,boxrule=1pt,fontupper=\ttfamily]
    You are an advanced reasoning agent that aims at improving reasoning through self refection. You were tasked to give a valid plan to a travel query given its relevant information and an automatic cost calculation environment. Only the selection whose name and city match the given information will be calculated correctly. You have tried to solve this task multiple times, and you will be given a reasoning path of one previous trials with their final results (either success or fail). Each reasoning path consists of a sequence of thought, action and observation at each timestep. Now, you need to reflect on the reasoning path of the current trial by classifying the error type for each action step and analyze whether each action step leads to a successful final plan. \\

    Step-by-step Instruction:\\
    1. Read all information and analyze the current trial. \\
    2. For the current step in the current trial, examine if it's valid and correct by checking whether it satisfies the given information and commonsense. Additionally, you need to compare it with the provided successful plan that satisfies the query to analyze the error. Then based on the given error taxonomy and the rationale on how the error taxonomy is proposed from all trials to the query, classify any potential errors by giving  corresponding error types if it contains errors. You can only select the error types from the error taxonomy and you  can leave it empty if there is no error. \\
    3. Give a critique as your explanation of each error classification for the current step in the current trial by considering whether the current step contributes to a valid final plan. Do not refer to other trials in your critique. \\
    4. Generate your output in a JSON format as a list of dictionaries. If there is no error, such as for successful trials, you can leave the list empty. Each dictionary item contains the following keys:\\
    \quad a. "error$\_$type": one error type for the current step. It must be from the given error taxonomy.\\
    \quad b. "critique": the critique of the current step for the classified error type.\\

    Given Information: \$ \{text\} \\
    
    Query: \$ \{query\} \\
    
    Error Taxonomy: \\
    \$ \{error\_taxonomy\}
    Error Taxonomy Rationale: \\
    \$ \{rationale\}\\
    
    \$ \{trial\_id\}: \\
    \$ \{current\_trial\} \\
    
    Valid Plan: \\
    \$ \{annotated\_plan\} \\

    Current Step: \\
    \$ \{current\_step\}
    \end{tcolorbox}
\caption{The designed prompt for constructing the error classification during the Intra-Level Learning.}
\label{fig:error_classification_prompt}
\end{figure*}

\begin{figure*}[ht]
    \centering
    \begin{tcolorbox}[left=5pt,right=5pt,colback=white,colframe=black,boxrule=1pt,fontupper=\ttfamily]
    You are an advanced reasoning agent that aims at improving reasoning through self refection. You were tasked to give a valid plan to each travel query given its relevant information and an automatic cost calculation environment. You have tried to solve multiple queries and made different errors in your trajectories that led to the failures of tasks. Now, you will be given multiple actions that you made with the same classified error type from different trials and each of these action has a corresponding critique regarding the error type. Your task is to give a comprehensive and high level diagnosis for the failure across different trials, and devise a new, concise, high level plan that aims to mitigate the same error. Use complete sentences.\\

    Error Type: \$\{error\_type\} \\

    \$ \{error\_trajectories\}
    \end{tcolorbox}
\caption{The designed prompt for the Inter-Level Learning.}
\label{fig:inter_prompt}
\end{figure*}

\section{Qualitative Example on TravelPlanner}
\label{sec:qualitative}
\begin{figure*}[ht]
\centering
\includegraphics[width=0.9\linewidth]{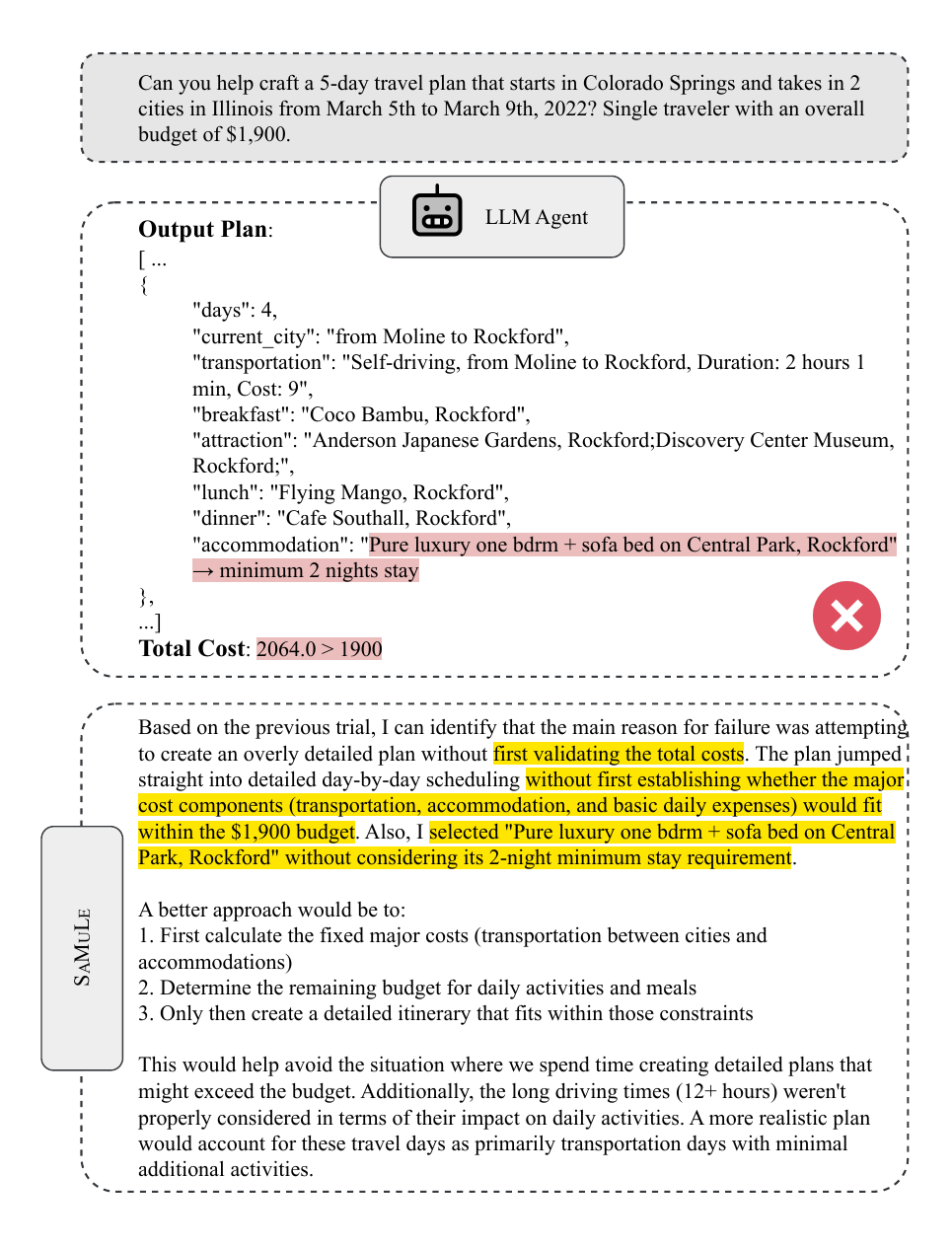}
    \caption{An example of the generated reflection by our approach on TravelPlanner.} 
    \label{fig:travel_example}
\vspace{-0.5cm}
\end{figure*}

\begin{figure*}[ht]
\centering
\includegraphics[width=0.9\linewidth]{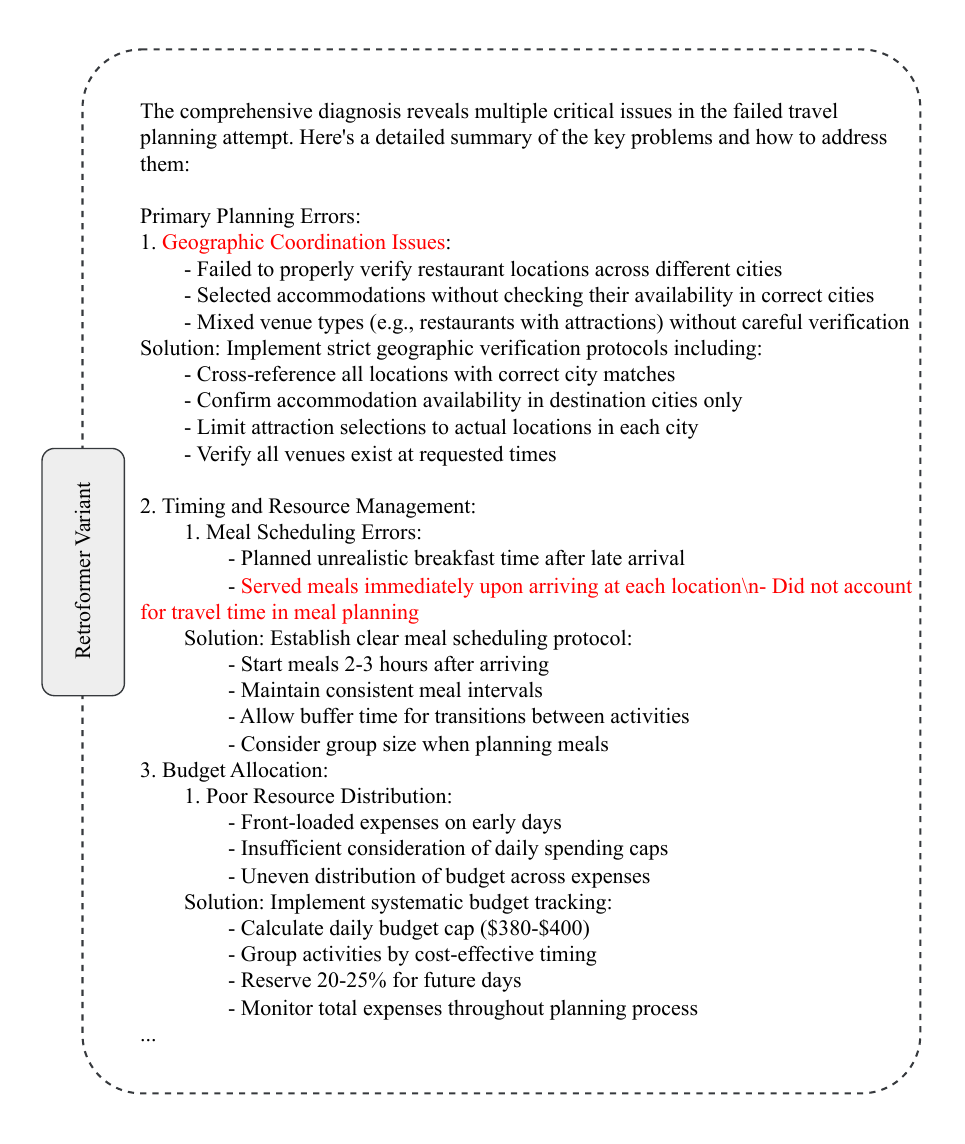}
    \caption{Part of the generated reflection by Retroformer Variant to the same input in Figure \ref{fig:travel_example}.} 
    \label{fig:travel_retroformer}
\vspace{-0.5cm}
\end{figure*}

We include a qualitative example of our approach on TravelPlanner in Figure \ref{fig:travel_example} and the output from Retroformer Variant in Figure \ref{fig:travel_retroformer}. Due to the limited space, we only show part of the output. In the output plan by the LLM agent, it violates two constraints: it selected a wrong accommodation which requires minimum 2 nights stay but only scheduled to stay for 1 night; the total cost of the plan is \$2064 and exceeds the budget \$ 1900. Then in the generated reflection by our approach, it pointed out these two errors and leads the agent to fix it in the next trial. However, Retroformer Variant identifies wrong errors, including geographic coordination issues and meal scheduling errors, as the geographic and meal timing information are not specified in the agent's output plan.

\section{Example of Generated Error Taxonomy}
\label{sec:error_taxonomy}
We show an example of the constructed error taxonomy during the meso-level of our Multi-level Reflection Synthesis on TravelPlanner in Figure \ref{fig:error_taxonomy}.
\begin{figure*}
    \centering
    \begin{tcolorbox}[left=5pt,right=15pt,colback=white,colframe=black,boxrule=1pt,fontupper=\ttfamily]
    Location Verification Error  \\
Accommodation Child Policy Violation \\ 
Accommodation Selection Error  \\
Travel Time Scheduling Error  \\
Accommodation Minimum Stay Violation  \\
Travel Time Planning Error  \\
Accommodation Pet Policy Violation  \\
Inefficient Resource Allocation  \\
Geographic Data Misinterpretation  \\
Restaurant Timing Error  \\
Budget Allocation Error  \\
Accommodation Smoking Policy Violation  \\
Transportation Planning Error  \\
Insufficient Data Recognition  \\
Invalid Location Selection  \\
Budget Constraint Violation  \\
Accommodation Capacity Violation  \\
Query Requirement Mismatch  \\
Attraction Planning Error  \\
Transportation Cost Error  \\
Incomplete Day Planning  \\
Restaurant Selection Error  \\
House Rules Violation  \\
Flight Connection Error  \\
Attraction Distribution Error  \\
Accommodation Party Policy Violation
    \end{tcolorbox}
\caption{The constructed error taxonomy on TravelPlanner.}
\label{fig:error_taxonomy}
\end{figure*}

\section{Generated Insights by Expel}
\label{sec:expel}
We present the insights generated by Expel on TravelPlanner in Figure~\ref{fig:expel_insights_travel}, and those on NATURAL PLAN in Figures~\ref{fig:expel_insights_natural_trip} and~\ref{fig:expel_insights_natural_meeting}. Overall, we observe that the insights produced for TravelPlanner are overly general, primarily focusing on cost-related issues while overlooking other critical error types, such as accommodation policy violations and travel time planning errors. This can lead the agent to have bias towards cost-related analysis in its future trials and ignore other errors, which further hurts the agent performance and even leads the pass rate on TravelPlanner to 0\%. Such a limitation is likely a result of Expel’s strict reliance on learning from successful trajectories, which are scarce in complex environments where agents struggle to produce successful plans. 
\begin{figure*}
    \centering
    \begin{tcolorbox}[left=5pt,right=5pt,colback=white,colframe=black,boxrule=1pt,fontupper=\ttfamily]
    - Calculate the total cost of the entire trip after planning each day to ensure it stays within the given budget, making adjustments as necessary. \{4\}\\
    - Allocate the budget strategically across all days of the trip, considering factors such as transportation costs, accommodation prices, and planned activities for each day. \{4\}\\
    - Start by outlining the major components of the trip that fit the given constraints before diving into detailed planning. \{3\}\\
    - When the total cost exceeds the budget, prioritize essential elements of the trip and look for cost-saving alternatives in transportation, accommodation, or activities to bring the plan within budget constraints. \{3\}\\
    - When encountering an error in planning (e.g., invalid dinner option), quickly adjust and retry with a valid alternative. \{2\}\\
    - Use the CostEnquiry action frequently to get real-time cost updates for each day's plan, allowing for immediate adjustments if necessary. \{2\}
    \end{tcolorbox}
\caption{The generated insights by Expel on TravelPlanner.}
\label{fig:expel_insights_travel}
\end{figure*}

\begin{figure*}
    \centering
    \begin{tcolorbox}[left=5pt,right=5pt,colback=white,colframe=black,boxrule=1pt,fontupper=\ttfamily]
    - Always start by clearly defining the travel query parameters, including origin, destination, dates, and any specific requirements. \{2\} \\
    - Utilize the automatic cost calculation environment efficiently by inputting all relevant information accurately and completely. \{2\} \\
    - Consider multiple travel options and compare their costs, duration, and convenience before recommending a plan. {2} \\
    - When presenting a travel plan, provide a clear breakdown of costs, itinerary details, and any important notes or restrictions. \{2\}
    \end{tcolorbox}
\caption{The generated insights by Expel on NATURAL PLAN (Trip).}
\label{fig:expel_insights_natural_trip}
\end{figure*}

\begin{figure*}
    \centering
    \begin{tcolorbox}[left=5pt,right=5pt,colback=white,colframe=black,boxrule=1pt,fontupper=\ttfamily]
    - Begin by gathering all essential meeting parameters, including participants, preferred dates and times, duration, and meeting objectives. \{2\} \\
    - Ensure all participants' availability is accurately collected to avoid scheduling conflicts. Leverage shared calendars or scheduling tools when possible. \{2\} \\
    - Consider alternative time slots and meeting formats (virtual, in-person, hybrid) to accommodate diverse participant preferences and logistical constraints. \{2\} \\
    - When proposing a meeting schedule, clearly outline the agenda, expected duration, and any required preparation to ensure effective participation. \{1\}
    \end{tcolorbox}
\caption{The generated insights by Expel on NATURAL PLAN (Trip).}
\label{fig:expel_insights_natural_meeting}
\end{figure*}

\end{document}